\documentclass[10pt,twocolumn,letterpaper]{article}

\usepackage{cvpr}
\usepackage{times}
\usepackage{epsfig}
\usepackage{graphicx}
\usepackage{amsmath}
\usepackage{amssymb}
\usepackage{bbm}

\usepackage[pagebackref=true,breaklinks=true,letterpaper=true,colorlinks,bookmarks=false]{hyperref}

\cvprfinalcopy 

\def\cvprPaperID{****} 

\ifcvprfinal\fi
\begin{document}

\title{Simultaneous $x, y$ Pixel Estimation and Feature Extraction for Multiple Small Objects in a Scene: A Description of the ALIEN Network}

\author{Seth Zuckerman\\
Aret\'e Associates\\
{\tt\small szuckerman@arete.com}
\and
Timothy Klein\\
Aret\'e Associates\\
{\tt\small tklein@arete.com}
\and
Alexander Boxer\\
Aret\'e Associates\\
{\tt\small aboxer@arete.com}
\and
Brian Lang\\
Aret\'e Associates\\
{\tt\small blang@arete.com}
\and
Christopher Goldman\\
Neurocore AI\\
{\tt\small cgoldman@neurocoreai.com}
}

\maketitle

\begin{abstract}
We present a deep-learning network that detects multiple small objects (hundreds to thousands) in a scene while simultaneously estimating their x,y pixel locations together with a characteristic feature-set (for instance, target orientation and color).  All estimations are performed in a single, forward pass which makes implementing the network fast and efficient. In this paper, we describe the architecture of our network --- nicknamed {\em ALIEN} --- and detail its performance when applied to vehicle detection. 

\end{abstract}

\section{Introduction}

\begin{figure*}[t]
\begin{center}
\includegraphics[width=0.95\linewidth]{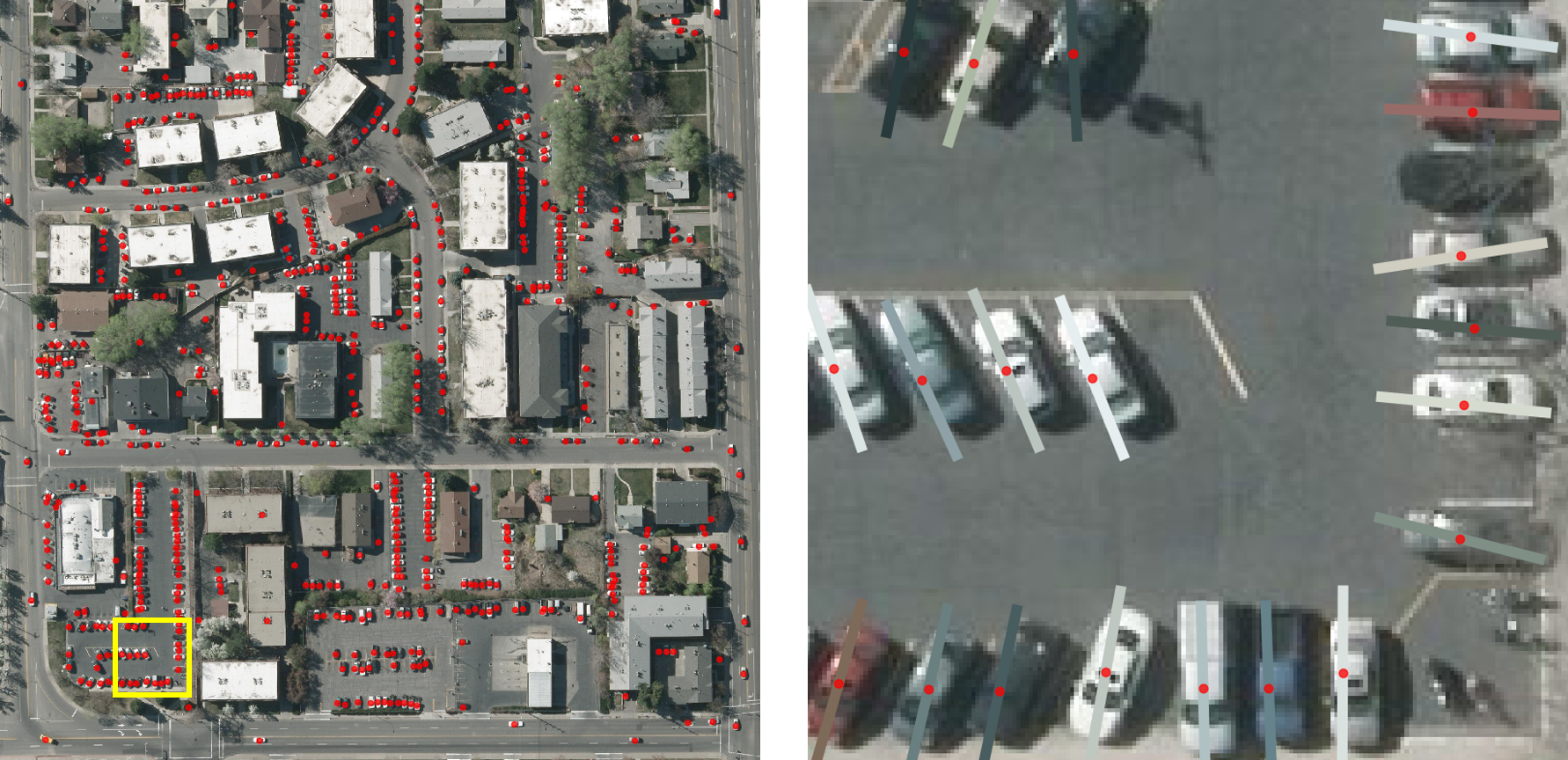}
\end{center}
   \caption{ALIEN network car detections in a megapixel holdout image (left). Detail showing inferences for orientation and color (right).}
\label{fig:cars}
\end{figure*}

A common task when analyzing imagery data is to detect, localize, and characterize a large and variable number of targets in a scene.    As a specific example (and to anticipate the data we will examine in this paper), suppose that you have aerial imagery of cities and within each image you want to (1.) count the total number of cars in the scene, which may range from zero to several thousand; (2.) obtain a list of the $x,y$ pixel locations of each car; and (3.) extract a set of characteristic features for each car, such as their color and orientation.    

Deep learning algorithms have been successful in each of these subtasks;  however, we are not aware of any network  that performs all of these subtasks simultaneously.  Through the novel use of masking functions to remove non-target inferences from the overall network loss, we have developed and trained a deep-learning network that performs each of these operations in a single forward pass.  In this paper, we describe the basic architecture of this network and summarize its performance when applied to publicly available datasets such as Cars Overhead with Context (COWC) \cite{mundhenk2016}.  

Insofar as deep-learning algorithms are more easily remembered when they have a catchy name, we have taken to calling ours the ALIEN (Attribute Localization and Instance Extraction Neural) network.  Figure \ref{fig:cars} shows the results of the ALIEN network when applied to a megapixel holdout image from the COWC dataset.

\section{Background}
Object detection in aerial imagery has been a persistent area of research, especially as it has important applications in fields ranging from urban planning and traffic monitoring \cite{palubinskas2008,xu2017} to defense \cite{sakla2017}.  Some notable, early efforts included Bayesian network, knowledge modeling, and histogram of oriented gradients approaches, while recent work has emphasized neural networks \cite{cheng2012, ruskone1996, gleason2011}. 

The goal of object detection is to locate one or multiple instances of an object in an image, rather than simply labeling the most prominent object in the scene. Regarding the recent neural network approaches to this problem, Du Terrail and Jurie \cite{duterrail2017} have argued that these fall into three, broad groups:

\begin{description}
	\item[Group 1. Sliding Window:] Windowing methods evaluate all parts of an image through various combinations of sub-windows. Neural network applications of this sort, which have been used since the 1990s \cite{matan1992}, became much more efficient with the advent of convolutional architectures \cite{sermanet2013}.  
	\item[Group 2. Region Proposal:] Instead of exhaustively evaluating all possible windows, region proposal methods attempt to gain efficiency by only evaluating those regions which are assessed to likely contain an object of interest. The R-CNN architecture described in Girshick \textit{et al.} \cite{girshick2014}, and its follow-on Fast R-CNN \cite{girshick2015}, is representative of this approach. The Faster R-CNN architecture of Ren \textit{et al.} \cite{ren2015} improves performance by replacing the hand engineered region proposal procedure with a learned region proposal network.
	\item[Group 3. Grid-Based Regression:] In contrast to the first two groups, grid-based methods recast the object detection problem as a regression. The YOLO (You Only Look Once) and follow-on YOLO9000 architectures are two well-known examples; these train a neural network to predict the location and bounding-box of an object with respect to a user defined grid \cite{redmon2016,redmon2016b}. 
\end{description}

\section{The ALIEN Network}

The ALIEN network is designed to detect, localize, and characterize objects within scenes that contain a large number (typically hundreds) of relatively small-sized targets.  By contrast, the object detection methods described above were designed  primarily to detect a small number (typically less than ten) of relatively large-sized targets.  The challenges of applying large-object detection techniques to the small-object regime have been noted, in particular, in Eggert \textit{et al.} \cite{eggert2017} and Sommer \textit{et al.} \cite{sommer2016}.  

The ALIEN network addresses these challenges with a novel algorithm.  Overall, it can be classed as a grid-based regression approach, albeit one that swaps bounding-box localizations for direct, $x, y$ pixel estimations.  However, what chiefly sets ALIEN apart is its use of masking functions; these facilitate multiple and simultaneous inferences since only those inferences which correspond to true targets are included in the overall network loss.  How this works is described below.

\subsection{Overview of the Approach}

The ALIEN network operates on image-cells of a pre-assigned pixel dimension.  Within each cell, a fixed number of {\it anchor-points} are arrayed, representing the maximum number of targets possible within a cell.  Inferences are made at each anchor-point which, in effect, transforms the problem from one where the total number of possible targets is variable to one where the total number of possible targets is fixed.  However, the first inference at each anchor-point is the probability that there is (or is not) a target present in the vicinity of that anchor-point.  In this way, and by using various masks, only the inferences that correspond to anchor-points where a target is present in the truth-data are included in the total loss and, ultimately, the final output.  The following subsections describe this procedure in more detail.         

\subsection{Anchor-Points}

\begin{figure}[t]
\begin{center}
\includegraphics[width=0.8\linewidth]{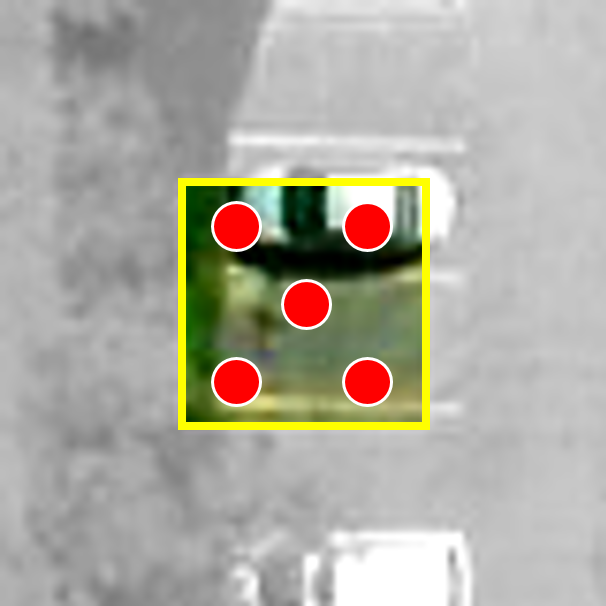}
\end{center}
   \caption{The ALIEN network is trained with chips, cells, and anchor-points. The network makes inferences at each anchor-point (red circles) within a cell (yellow square).  The number of anchor-points corresponds to the maximum number of targets that can fit inside of a cell.  The larger chip (black and white) provides a contextual border around the cell, but targets outside the cell are not included in the truth-data associated with the chip.}
\label{fig:anchorpoints}
\end{figure}

To make the anchor-point concept more concrete, consider a specific example.  Images from the COWC dataset can range in dimension up to several thousand pixels per side.  Conveniently, however, all of the images were taken with a consistent GSD (Ground Sample Distance) such that all the cars from one image to the next are approximately the same size (about 30 pixels long and 10 pixels wide).  

The basic image-unit used in our implementation of the ALIEN network was a cell of dimensions 32$\times$32 pixels.  Five anchor-points were positioned in each cell and arranged in the quincunx pattern of a standard playing die (Figure \ref{fig:anchorpoints}).  While the choice of these values (cells of 32$\times$32 pixels  and five anchor-points per cell) is, in principle, arbitrary, their relationship is constrained to ensure that there cannot be more targets per cell than anchor-points. In other words, in the COWC dataset, it is not possible to find more than five cars in a single, 32$\times$32 pixel region. 

As a practical matter, overall performance is significantly improved when the network is trained on image-{\em chips} that are somewhat larger than an image-cell.  The effect of this is to provide a contextual border which helps the network learn to identify targets even when they are only partially inside of a cell (a common occurrence).  To implement this, our network was trained on 80$\times$80 pixel chips; however, the truth-data associated with each chip corresponded only to the 32$\times$32 pixel cell positioned at the chip's center.  In other words, if a car was present in a chip, but not the cell, there would be no record of that car in the chip's truth-data.  The relationship between chip, cell, and anchor-point is illustrated in Figure \ref{fig:anchorpoints}.

\subsection{Existence and Loss}

\begin{figure*}
\begin{center}
\includegraphics[width=0.85\linewidth]{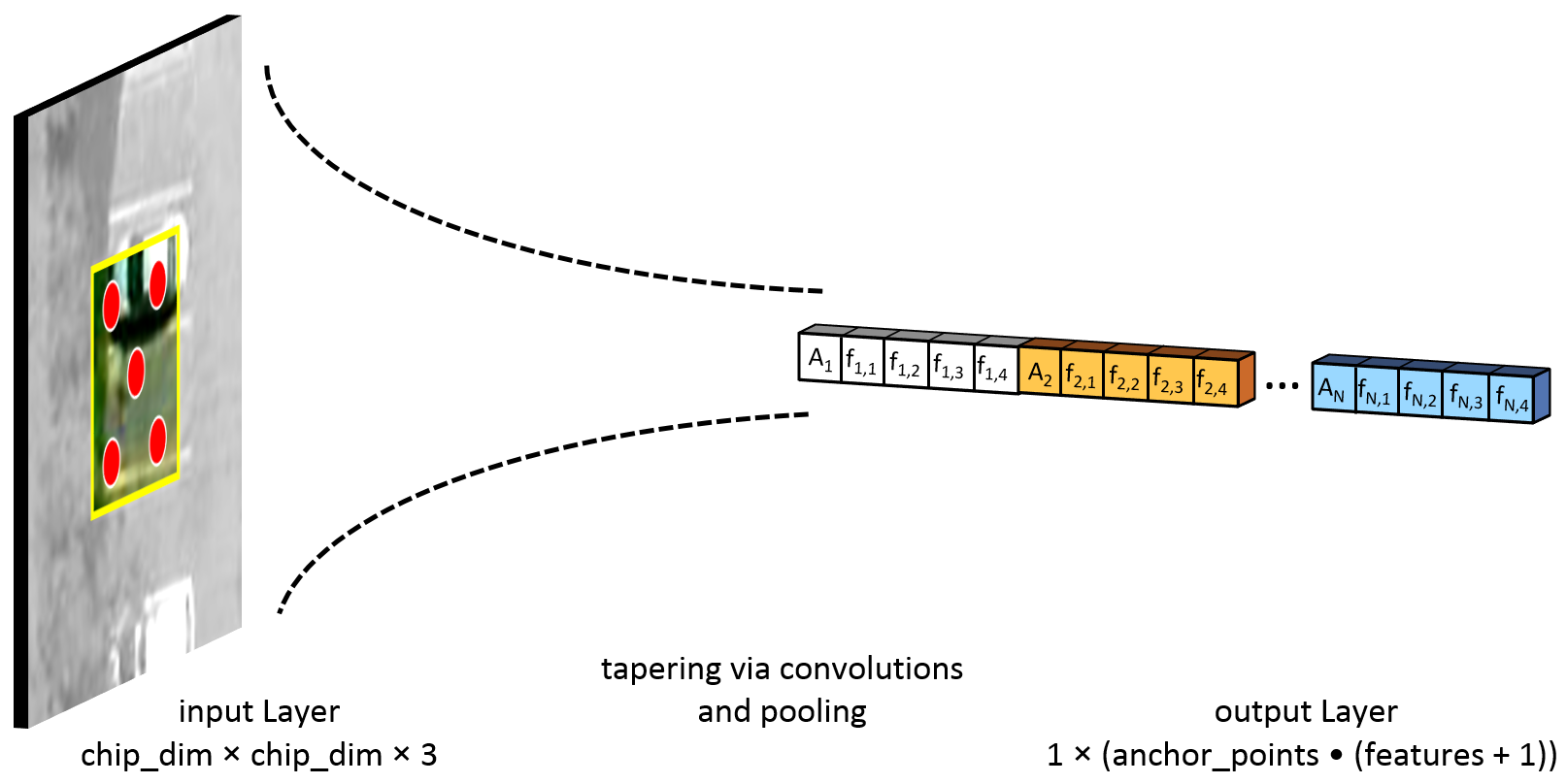}
\end{center}
   \caption{Conceptual diagram of the ALIEN network architecture. The inferences made at each anchor-point are grouped by color.}
\label{fig:alien}
\end{figure*}

The underlying stratagem of the ALIEN algorithm is that inferences are made at {\em every} anchor-point, regardless of whether a target is present or not. However, the contribution of any individual inference to the total network loss will depend on whether a target is, or is not, present in the truth-data associated with that anchor-point.  

This procedure can be expressed formally by using an indicator function, denoted as $\mathbbm{1}\{\cdot\}$, which evaluates to either 1 or 0 depending on the truth-status of the expression in the brackets.  The total network loss, $J$, can therefore be written very generally as follows:

\begin{equation}
J = \sum_{m,n}\lambda_{m}\cdot f(\mathbbm{1}\{y^{(0,n)}\})\cdot g(y^{(m,n)}, \hat{y}^{(m,n)})
\label{eq:loss}
\end{equation}

Equation \ref{eq:loss} describes a network that makes $M$ inferences at $N$ anchor-points.  The zero-indexed inference, $ \hat{y}^{(0,n)}$, is the estimated probability that a target exists at anchor-point $n$;  the remaining inferences, $\hat{y}^{(m \neq 0,n)}$, pertain to additional features (\eg, color) associated with the target.  

The contribution to the total loss, $J$, from the inference for feature $m$ at anchor-point $n$ contains a component, $g$, which is the contribution made under the assumption that a target exists at anchor-point $n$.  Thus, for a feature presumed to be normally distributed about its mean (\eg, the $x$ or $y$ pixel offset), the component $g$ would simply be the mean-squared error that takes, as its arguments, the network inference, $\hat{y}^{(m,n)}$, and its corresponding truth,  $y^{(m,n)}$.   

The decision whether, or to what extent, to include the contribution $g$ in the total loss is governed by a second function, $f$, whose sole argument is $\mathbbm{1}\{y^{(0,n)}\}$, namely, the truth-status of the existence of a target at anchor-point $n$.  Since inferences of certain features (\eg, target-color) are meaningless for non-existent targets, $f$ allows such inferences to be excised altogether from the total loss.  

For the most part, $f$ is simply the identity function, which gives the indicator function $\mathbbm{1}\{y^{(0,n)}\}$ full control to switch $g$ on or off as appropriate.  However, the zero-indexed feature --- target existence --- is categorical (target, no-target) and its contributions to the overall loss are computed as a standard, binary cross-entropy.  Here the contributions from non-targets are, of course, every bit as important as the contributions from targets, which makes it necessary to retain $f$ in Eq.~\ref{eq:loss}.      

The last component to the total network loss are weights $\lambda_m$ which determine the relative magnitudes of the contributions from each of the $m$ features.  While the question of how to optimally combine multiple losses in a deep-learning network is, theoretically, a subtle one, we found that our network was surprisingly robust to the choice of these weights.  Provided that the contributions to the loss from each feature were within an order of magnitude or two, the newtork training times and performance hardly varied.  

Nor did we encounter any practical limitations to the number of features that can be included in a loss function of the form of Eq.~\ref{eq:loss}.  The ALIEN implementation discussed in this paper made 8 simultaneous feature inferences, in addition to a target/no-target inference, at every anchor-point.  These were: $x$ and $y$ pixel-offsets of the target from each anchor-point (2 features); target-color, estimated as sine and cosine components for hue, and scalar components for saturation and value (4 features); and target-orientation, estimated as a sine and cosine pair (2 features).   

Creating a working implementation of Eq.~\ref{eq:loss} proved reasonably straightforward using the Keras deep-learning library with a TensorFlow backend.  Keras offers a wide leeway for defining customized loss-functions and, crucially, the \texttt{tf.boolean\_mask} operation can be used as a versatile, one-line implementation of the indicator function $\mathbbm{1}\{y^{(0,n)}\}$.      

\subsection{Network Architecture}

From a network architecture perspective, the job of the ALIEN network is to transform its input data, in the form of RGB imagery, into a one-dimensional vector of inferences.  This output layer will have a length equal to (the number of anchor-points per cell) $\times$ (the total number of estimated target-features plus one).  The plus-one is the zero-indexed `feature' representing the probability that a target exists, or does not exist, in the vicinity of that anchor-point.  

The ALIEN network makes the conversion from a 3-dimensional, RGB input-layer to a one-dimensional output layer in a straightforward way via successive two-dimensional convolutions and poolings.  Figure \ref{fig:alien} shows a conceptual diagram of this process.   

Our implementation of the ALIEN architecture was constructed as a sequential model of 14 two-dimensional convolutions, 4 max-pooling layers, 3 dropout layers, and a final sigmoid output.    Each convolutional output has a leaky ReLU activation ($\alpha = 0.1)$, and the initial convolutions are applied with valid padding in order to contribute to dimensionality-reduction. 8 plus 1 inferences were made at each of the 5 anchor-points per cell resulting in a total of 45 inferences for every 80$\times$80 pixel image-chip. Table \ref{tab:summary} shows a model summary of the layers of our ALIEN network.

\begin{table}
\begin{center}
\small
\begin{tabular}{|c|c|c|c|}
\hline
Layer Type & Output Shape & Filter Size & Channels \\
\hline\hline

Input & $80\times80\times3$ & & \\
\hline

Convolution 2D & $78\times78\times16$ & $3\times3$ & $16$ \\
\hline

Convolution 2D & $76\times76\times16$ & $3\times3$ & $16$ \\
\hline

MaxPool 2D & $38\times38\times16$ &  &  \\
\hline

Convolution 2D & $34\times34\times16$ & $5\times5$ & $16$ \\
\hline

Convolution 2D & $32\times32\times16$ & $3\times3$ & $16$ \\
\hline

Convolution 2D & $28\times28\times16$ & $5\times5$ & $16$ \\
\hline

MaxPool 2D & $14\times14\times16$ &  &  \\
\hline

Convolution 2D & $12\times12\times32$ & $3\times3$ & $32$ \\
\hline

Convolution 2D & $8\times8\times32$ & $5\times5$ & $32$ \\
\hline

MaxPool 2D & $4\times4\times32$ &  &  \\
\hline

Convolution 2D & $2\times2\times32$ & $3\times3$ & $32$ \\
\hline

MaxPool 2D & $1\times1\times32$ &  &  \\
\hline

Convolution 2D & $1\times1\times256$ & $1\times1$ & $256$ \\
\hline

Dropout (10\%) & $1\times1\times256$ &  &  \\
\hline

Convolution 2D & $1\times1\times256$ & $1\times1$ & $256$ \\
\hline

Dropout (10\%) & $1\times1\times256$ &  &  \\
\hline

Convolution 2D & $1\times1\times256$ & $1\times1$ & $256$ \\
\hline

Dropout (10\%) & $1\times1\times256$ &  &  \\
\hline

Convolution 2D & $1\times1\times256$ & $1\times1$ & $256$ \\
\hline

Convolution 2D & $1\times1\times256$ & $1\times1$ & $256$ \\
\hline

Convolution 2D & $1\times1\times45$ & $1\times1$ & $45$ \\
\hline

Sigmoid & $1\times1\times45$ &  &  \\
\hline

\end{tabular}
\end{center}

\caption{ALIEN model summary}\label{tab:summary} 
\end{table}

The main subtlety in implementing the ALIEN architecture is to ensure that the truth-data on which the network is trained supports the ALIEN algorithm.  This entails ensuring that any targets within a cell are assigned to anchor-points in a systematic and unambiguous way.  Since two or more targets can be close to the same anchor point, their assignment is best handled recursively. Specifically, the target closest to any of the anchor-points gets assigned first, which then eliminates that anchor-point for the remaining targets, \etc  A method like this avoids the trap where  the first listed target is always assigned to anchor-point 1, the second listed target to anchor-point 2, and so on, which has the effect of obliterating the spatial information encoded by the anchor-points themselves.

\subsection{A Fully Convolutional Implementation}

The ALIEN architecture just described is tailored for training a network on $80\times80$ pixel chips.  However, once the weights have been trained, the ALIEN network is capable of evaluating images of any size.  This is accomplished by running the saved weights convolutionally over the larger image.  

This `fully convolutional' implementation generates inferences over multiple $80\times80$ pixel regions whose spacings are determined by the convolution's stride.  Recall that while the ALIEN network is trained to treat regions the size of a chip, it only generates inferences on the smaller, $32\times32$ pixel cell at the center of the chip.  Thus, in order for the fully convolutional implementation to achieve full coverage, the convolutional stride must be set equal to the cell-size.

The outputs of this process will be a long list of inferences totaling (the number of anchor-points per cell) $\times$ (the total number of features plus one) $\times$ (the number of cells required to cover the evaluation image).  However, so long as some careful bookkeeping is employed to reconstruct the addressing of the outputs, there is nothing that prevents the ALIEN network from  running on megapixel images just as effectively as it can on small chips.  And since all inferences are made simultaneously in a single forward pass, this makes the ALIEN network an especially fast and powerful algorithm for performing the types of large-scale scene analyses that are either tedious, or simply infeasible, with legacy algorithms. For instance, we were able to process scenes as large as 100 megapixels without ever needing to subdivide the original image into smaller chips.

\section{Application to the Cars Overhead with Context (COWC) Dataset}

\begin{figure}[t]
\begin{center}
\includegraphics[width=0.9\linewidth]{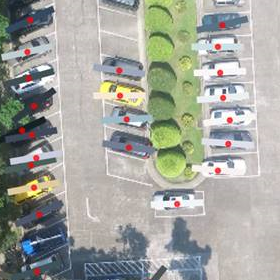}
\end{center}
   \caption{The ALIEN network was similarly successful when applied to imagery from the drone-based CARPK dataset once the GSD was made comparable to the training data.}
\label{fig:cars2}
\end{figure}

The primary dataset used to develop, train, and test the ALIEN network was the Cars Overhead with Context (COWC) dataset \cite{mundhenk2016}.  This is a publicly available amalgamation of six, separate aerial datasets from distinct geographical locations.  The combined set consists of 32 very large (megapixel) images containing, in total, over 32,000 unique cars.  Each large image comes with a corresponding pixel-mask annotation, where single hot pixels indicate the approximate center of each car.  

Of these 32 annotated images, 29 were used as a training set and 3 were used as a validation set.  Additionally, a separate set of 20, unannotated images is provided as a holdout set.  Figure \ref{fig:cars} shows a typical result of the ALIEN network on a $2048\times2048$ pixel image from the holdout set.  The entire scene contains 625 cars; the ALIEN network correctly detected 616 of these with 43 false alarms, yielding an F1 classification score of 96\%. 

The ALIEN detections and localizations are depicted in Fig.~\ref{fig:cars} as red dots superimposed on each target's inferred location (left).  The region enclosed by the small yellow box, is magnified (right) to show the ALIEN network's inferences for target orientation and target color.  The inferred orientations are indicated by the orientations of the solid line segments; the color of these segments indicates the inferred target hue.  

The performance of the ALIEN network averaged over the entire COWC holdout set is listed in Table \ref{tab:results}.  The mean-squared pixel-error of 2.5 pixels is well within the dimensions of the targets.

\begin{table}
\begin{center}
\begin{tabular}{|l|r|}
\hline
Metric & Performance \\
\hline\hline
True Detections & 95\% \\
False Alarms & 6\% \\
Localization Accuracy & 2.5 px \\
Hue Accuracy & 25$^\circ$ \\
Orientation Accuracy & 9$^\circ$ \\
\hline
\end{tabular}
\end{center}
\caption{ALIEN network performance on the COWC holdout set} \label{tab:results} 
\end{table}

\subsection{Additional Imagery}

As an additional test, we examined images from an entirely separate database: the drone-based car parking lot dataset (CARPK)  \cite{Hsieh2017}.  These images have a different GSD than the COWC images, but once this discrepancy was addressed through resampling, the results were similarly successful (Figure \ref{fig:cars2}).

\section{Conclusion}

The ALIEN network has demonstrated that it can be successfully applied to a problem that is frequently encountered in aerial image analysis, namely, the detection, localization, and characterization of multiple small objects in a scene.  Moreover, by employing masking functions when computing the overall network loss, each of these subtasks can be performed simultaneously in a single forward pass.  There is no limit to the number of features that can be simultaneously estimated this way which makes the ALIEN architecture particularly versatile.  Lastly, the ALIEN network is designed to run convolutionally over images of any size, such that even very large (megapixel) images can be processed in one pass without the need for additional chipping or scaling.

\end{document}